# History Playground: A Tool for Discovering Temporal Trends in Massive Textual Corpora

*Thomas Lansdall-Welfare* and *Nello Cristianini*

Intelligent Systems Laboratory, University of Bristol, Bristol BS8 1UB, United Kingdom

## Abstract

Recent studies have shown that macroscopic patterns of continuity and change over the course of centuries can be detected through the analysis of time series extracted from massive textual corpora. Similar data-driven approaches have already revolutionised the natural sciences, and are widely believed to hold similar potential for the humanities and social sciences, driven by the mass-digitisation projects that are currently under way, and coupled with the ever-increasing number of documents which are "born digital". As such, new interactive tools are required to discover and extract macroscopic patterns from these vast quantities of textual data. Here we present History Playground, an interactive web-based tool for discovering trends in massive textual corpora. The tool makes use of scalable algorithms to first extract trends from textual corpora, before making them available for real-time search and discovery, presenting users with an interface to explore the data. Included in the tool are algorithms for standardization, regression, change-point detection in the relative frequencies of n-grams, multi-term indices and comparison of trends across different corpora.

## 1 Introduction

Massive textual corpora including newspapers archives, collections of webpages and social media posts contain information about many real world events (Leban, et al., 2014) (Leetaru & Schrodt, 2013), changes in popular opinion, mood and collective behaviour (Golder & Macy, 2011) (Dodds, et al., 2011) (Lansdall-Welfare, et al., 2012) (Lansdall-Welfare, et al., 2016), while also containing hidden patterns about a wide range of other concepts and topics (Michel, et al., 2011) (Lansdall-Welfare, et al., 2017) (Lampos, et al., 2010) (Caliskan, et al., 2017). As it stands, data-driven approaches are needed to analyse these data sets due to the infeasibility of more traditional close-reading techniques to cope with the sheer volume of information contained within such large numbers of texts.

Various studies in recent years (Michel, et al., 2011) (Lansdall-Welfare, et al., 2017) (Dexter, et al., 2017) have begun to demonstrate the potential of taking a data-driven approach, but are often limited to interdisciplinary teams, requiring significant programming knowledge and expertise in natural language programming, machine learning and big data, working alongside humanities scholars and social scientists.

The field of digital humanities would benefit greatly from a democratisation of tools, developed to enable scholars to more easily study the output of the numerous mass-digitisation projects which are currently taking place around the world, such as the Chronicling America Project (Library of Congress and National Endowment for the Humanities, 2017), the British Newspaper Archive (Findmypast Newspaper Archive Limited and the British Library, 2017) and Trove (National Library of Australia, 2017). The creation of the Google n-grams viewer to visualise relative frequency timelines for words in millions of books has greatly affected the field of digital humanities, inspiring over a thousand academic papers, and defined a new area of investigation; the data itself might have been available in other formats (Google books, Gutenberg project, etc.) but just the creation of a digital archive is not sufficient to analyse macroscopic trends. This is further explored in (Terras, et al., 2017), where the authors find *"that there are at present too many technical hurdles for most individuals in the arts and humanities to consider analysing large-scale open data sets"*. As the field progresses with the creation of digital newspaper archives, a new way to access that data is necessary, one that builds on the lessons of the books n-grams experience, and greatly extends them and applies them to these public resources, allowing researchers to interact online with those resources, and adds more data-analytic tools to the basic visualisation. Automation of data-driven techniques that allow for expert knowledge to be used in the study design and later analysis of the data are key at this stage to realise the potential of the field.

In this direction, we present an interactive web-based tool named History Playground, providing a way for users to explore, interrogate and generally discover macroscopic patterns and trends within massive textual corpora, leveraging the aid of automated data-driven methods to explore and play with the data.

## 2 Background and Context

Within our research group, we have completed numerous automated content analysis studies, working with a range of scholars from the humanities and social sciences (Lansdall-Welfare, et al., 2014) (Jia, et al., 2016) (Lansdall-Welfare, et al., 2017). During this time, we found that we needed a way to allow our traditionally qualitative collaborators to touch and play with the data, giving them a sense of what is possible using automated means, and where these types of results can be combined with close-reading approaches to attempt to bridge the gap between the two distinct methodologies.

The web-based tool was developed in an effort to bring the worlds of quantitative and qualitative analysis closer together, allowing us to explore a complementary approach of combining close reading with data-driven techniques. This has allowed us to build on the strengths of each type of approach when taken independently, tackling the drawbacks of each approach individually.

More specifically, the main critique we have experienced of quantitative, data-driven approaches in the digital humanities is that only a shallow understanding of the data is accessible, with a depth of meaning missing from the automated methods. Similarly, we find that qualitative methods are criticised for their comparatively tiny sample sizes (Manovich, 2011), which can easily miss data relevant to the study at hand, or that they are too closely focused on a very particular aspect of the data.

We therefore looked to create a tool that would find a complementary balance between these two extremes, where large sample sizes could be investigated while still enabling close-reading to infuse studies with a deeper understanding and nuance that can still be difficult to achieve when pursuing automated, data-driven approaches.

Internally within our group, the tool has been through many iterations, with various types of data being visualised and made interactive, from n-gram time series and sentiment analysis, to association networks and networks of subject-verb-object triplets, as shown in (Lansdall-Welfare, 2015, pp. 139-150). For example, in (Lansdall-Welfare, et al., 2014), we studied the impact of the Fukushima nuclear disaster on media coverage of nuclear power in over five million newspaper articles, using the tool as the main vehicle for discovering and visualising the results. This not only focused on an analysis of how the entities (people, organisations, locations) associated with nuclear power changed, but also on how the association with more specific science related topics, research institutions and diseases were affected. However, we found that many of these features need to be focused or customised within the context of a given study, which can be difficult to interpret or use when provided in a more general form.

Therefore, as a first step on a longer path to making useful tools and software accessible to other researchers, we feel that releasing a more general tool, similar in spirit to the Google Books Ngram Viewer[1], that over time can be updated to include a wide number of datasets using the same interface and methods is of greater benefit to the community than various customised versions for specific domains within the humanities. Our hope is that the History Playground helps to address the questions raised in (Cheney, 2013) of what tools exist and who will make them to allow researchers to access large scale corpora and visualise them, bypassing the *"restrictive interfaces of online archives"* that currently make digital humanities studies difficult (Nicholson, 2012).

---

[1] Google Books Ngram Viewer is available online at: https://books.google.com/ngrams

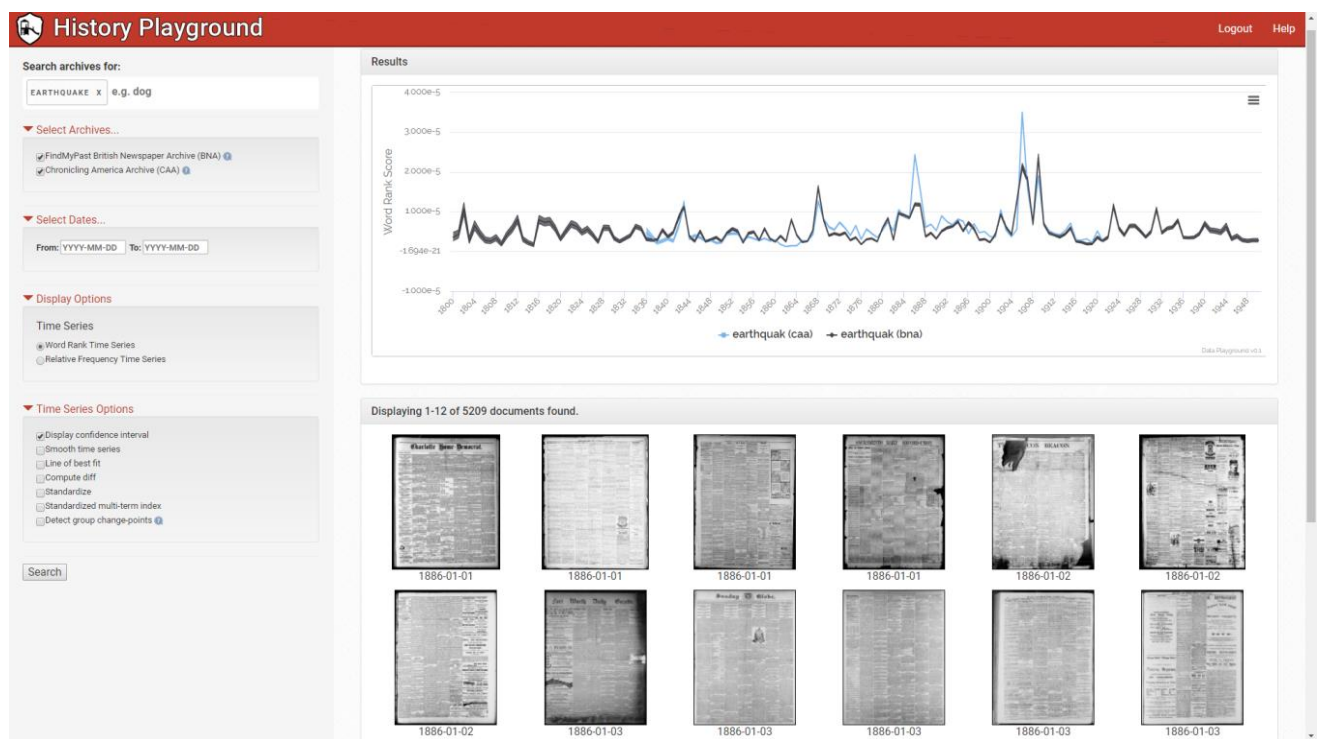

*Fig. 1 – Screenshot of the History Playground, available online at playground.enm.bris.ac.uk*

## 3 History Playground

The aim of the History Playground is to allow users to search for small sequences of words, otherwise referred to as n-grams, and retrieve their relative frequency over the course of history within a given corpus. Changes in the relative frequency of an expression can signal increased or decreased relevance of a topic, entity or event, or a change in public discourse, attitude or language. But detecting such a change is both technically difficult, without the right tools, and not sufficient if taken alone: further reading of the sources needs to be performed, to refine the queries. The interactive, web-based tool[2], is composed of a web interface where users can submit queries and interact with the resulting visualisation, a NoSQL database containing pre-computed n-gram time series, and an index containing further meta-data and the full-text of the documents in each corpus. A screenshot of the History Playground is depicted in Fig. 1.

The main visualisation of the tool is presented as a time series of the relative frequency of a given n-gram, which forms the start of any investigation into its changes and continuities over time. Clicking on a time series will present the user with the pages from the underlying corpus matching the given n-gram and date, allowing the user to further interrogate the corpus via close-reading. Beyond just displaying the n-gram time series, several different options are built in to the tool, enabling users to automatically compute different statistics of the data such as confidence intervals, standardisation, rates of change and change-points across multiple time series at once, as well as offering the ability to download the data from queries either as a figure, or in a format suitable for further analysis in statistical software packages. As a starting point, we are making accessible the time series for two large English-language corpora: the n-grams from the corpus of local British periodicals and newspapers between 1800-1950 examined in (Lansdall-Welfare, et al., 2017), and the n-grams available in the corpus released by the Chronicling American project (Library of Congress and National Endowment for the Humanities, 2017). For reference, details of how the n-grams are computed, along with further information on the different time series options, and how they are computed is included in each of the following sections, with the full processing pipeline for a given query illustrated in Fig. 2.

---

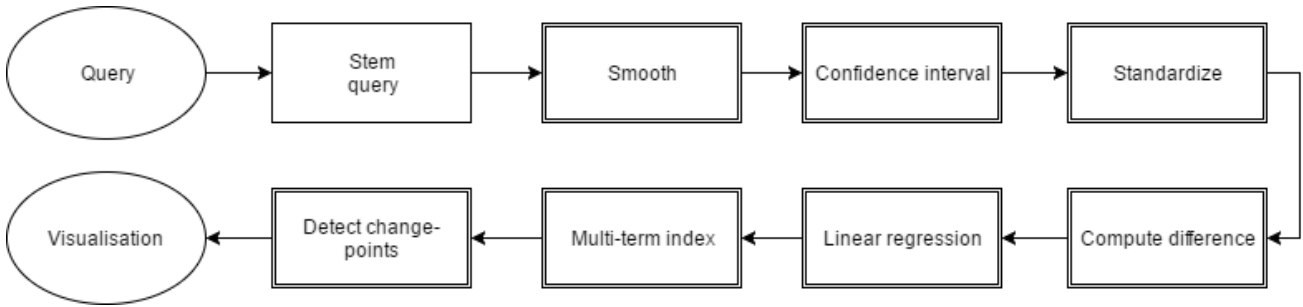

*Fig. 2 – Flowchart showing the ordering of the components in the query processing pipeline for each n-gram query submitted to the web interface. Double-bordered boxes are optional and can be trigger by selecting them in the interface.*

### 3.1 Relative frequency of n-grams

The relative frequency of an n-gram in a given corpus can be computed by counting how often in each document the n-gram occurs and then aggregating the counts into a total for each time interval based upon the publication date of the documents. This total for each n-gram is then divided by the sum over all n-grams in the time interval, giving the relative prevalence of each n-gram within the corpus at the time of its usage. Depending on the corpus, the relative frequency of an n-gram can be computed for different time scales, alternatively computing the time series at daily, weekly, monthly, yearly or any other appropriate interval. The relative frequency $f_w(t)$ of an n-gram $w$ at time $t$ can therefore be expressed as

$$f_w(t) = \frac{c_w(t)}{\sum_{i=0}^{V} c_i(t)} = \frac{c_w(t)}{n(t)}$$

where $c_w(t)$ is the count of how many times the n-gram $w$ occurs at time $t$, $n(t)$ is the sum count of n-grams seen at time $t$, and $V$ is the vocabulary of n-grams in the corpus, indexed by $i$.

However, due to the different properties of individual corpora, including the corpus size, language and size of the lexicon, we found that while the relative frequency of an n-gram works well for an individual corpus, they are often not directly comparable across corpora. In response to this, we developed the word rank score described in the following section as a different way to estimate the relative importance of an n-gram in a given corpus, that is closely related to the relative frequency, but allows for more direct comparisons between corpora of different size. This quantity gives us a more robust way to detect changes in the relative importance of expressions, over time.

### 3.2 Word rank scores for n-grams

The key observation behind our proposed n-gram scoring method, the word rank score, is that relative frequencies are affected by the size of the corpus, the size of the lexicon and also by the presence of OCR noise, and that this influence is stronger in the low-frequency end of the tail. This makes it very difficult to compare n-grams from different corpora to each other using their relative frequency. However, relative positions in the frequency ranking of words are much more robust to these effects, particularly at the more frequent end of the distribution, therefore we would like estimate changes in relative importance of a word by observing its change in the frequency ranking.

The work rank score for an n-gram is derived from Zipf's law (Zipf, 1949), which states that for natural languages, the frequency of any n-gram is inversely proportional to its rank by frequency. Using this law, we first assign a rank $k$ to each n-gram at each time point, based upon the total ordering of the relative frequencies, where n-grams with the same relative frequency are given the same rank. The most frequent n-gram is assigned rank $k = 1$, the second most frequent $k = 2$, and so on. By using the rank of the n-gram, rather than its observed relative frequency we can estimate the relative probability of a given n-gram in a more robust way, which enables fairer comparisons across different corpora.

More formally, using Zipf's law we approximate the relative frequency $f_w(t)$ using our word rank score $r_w(t)$, calculated as

$$f_w(t) \cong r_w(t) = \frac{1}{k_w(t)H_{n(t)}} - \varepsilon(t)$$

$$\varepsilon(t) = \frac{1}{k_\emptyset(t)H_{n(t)}}$$

where $k_w(t)$ is the rank $k$ of n-gram $w$ at time $t$, $H_{n(t)}$ is the generalised $n^{\text{th}}$ harmonic number, $n(t)$ is the sum total of n-grams seen at time $t$ and $\varepsilon(t)$ is a time dependent correction term that uses $k_\emptyset(t)$, the rank of the zero frequency n-grams at time $t$, to calibrate time series to the same baseline, e.g. $f_w(t) = 0$ for zero frequency n-grams at all values of $t$.

### 3.3 Smoothing of time series

The time series of an n-gram can often result in a signal with high variability or short term fluctuations, which can obscure the overall trajectory over time, making it difficult to assess the general pattern exhibited by the n-gram. One technique for extracting this general trend is to smooth the time series, removing the short-term fluctuations from the signal. The option to smooth the time series is implemented within the History Playground using a moving average, where the value at each time $f_w(t)$ is calculated as

$$f_w(t) = \frac{f_w\left(t - {}^\omega\!/_2\right) + \cdots + f_w(t) + \cdots + f_w\left(t + {}^\omega\!/_2\right)}{\omega}$$

where $\omega \in \{2k + 1 : k \in \mathbb{Z}\}$ is the smoothing window size, set by default to 3. Larger values of $\omega$ can be set individually per corpus and will smooth progressively longer term fluctuations in the time series.

### 3.4 Confidence intervals

When reading a time series for an n-gram, it is useful to know how confident one can be in the estimation of the relative frequency (or word rank score) at a given time point, enabling one to know if a change in a time series is due to noise or poor estimation, or if it is a real change in the signal. Depending on the size of the corpus being analysed, and the resolution at which one is looking, there may be too few instances of the n-gram, or too small a corpus, to draw any meaningful conclusions from the time series because the relative frequency can only be poorly estimated *i.e.* within a large range of values. Confidence intervals are therefore used to show the largest and smallest values a relative frequency or word rank score can be expected to take with 95% confidence. If the upper and lower confidence intervals for two time points do not overlap, then we know that these values can be considered significantly different. In order to help users know when this might be the case, we show the confidence interval for each n-gram time series.

We can calculate the confidence interval of a time series using the continuity-corrected Wilson interval, an approximation to the Binomial population confidence interval (Wallis, 2013). The upper and lower bounds ($f_w^+(t)$ and $f_w^-(t)$ respectively) for an estimate $f_w(t)$ at a given time point are calculated as

$$f_w^+(t) \equiv \min\left(1, \frac{2n(t)p(t) + z_{\alpha/2}^2 + \left\{z_{\alpha/2}\sqrt{z_{\alpha/2}^2 - \frac{1}{n(t)} + 4n(t)p(t)[1 - p(t)] - [4p(t) - 2] + 1}\right\}}{2[n(t) + z_{\alpha/2}^2]}\right),$$

$$f_w^-(t) \equiv \max\left(0, \frac{2n(t)p(t) + z_{\alpha/2}^2 - \left\{z_{\alpha/2}\sqrt{z_{\alpha/2}^2 - \frac{1}{n(t)} + 4n(t)p(t)[1 - p(t)] + [4p(t) - 2] + 1}\right\}}{2[n(t) + z_{\alpha/2}^2]}\right),$$

where $n(t)$ is the sum total of n-grams seen at time $t$, $c_w(t)$ is the given n-gram count, $a$ is the significance level (set to 0.05) and $z$ is corresponding standard normal deviate (set to 1.95996).

### 3.5 Linear Regression

Often one wishes to know whether a time series is increasing or decreasing, and if so, what the gradient of this change might be. For example, this allows one to know whether a series is increasing more quickly than another. In the tool, users can choose to display the line of best fit for a time series, with additional information on the gradient and intercept shown below the chart.

The line of best fit is calculated using linear least squares regression (Watson, 1967) as implemented in the least-squares JavaScript package (Richardson, 2016), along with the standard error, which denotes how well the data is modelled by the equation of the line found using regression.

### 3.6 Standardisation

The frequency of words within a corpus of natural language follow Zipf's law, which states, as previously mentioned, that the frequency of any word is inversely proportional to its statistical rank (Zipf, 1949). This simply means that the most frequent word will occur approximately twice as often as the next most frequent word, three times as often as the following next most frequent and so on.

As such, it can often be difficult to compare n-grams with largely different relative frequencies, due to the potentially orders of magnitude difference between them. However, if one is unconcerned with the magnitude of the relative frequency of a set of n-grams, and instead wish to compare other quantities, such as change-points or the overall trend, a simple way to achieve this is to standardise the time series.

Standardisation of a time series is a common technique, otherwise known as z-scoring, and involves subtracting the mean of the time series from each time point, and dividing through by the standard deviation. The resulting time series will have a zero mean and unit variance, allowing them to be compared on the same scale.

For each time interval in an n-gram time series $f_w(t)$, the standardized time series $z_w(t)$ is calculated as

$$z_w(t) = \frac{f_w(t) - \mu_w}{\sigma_w}$$

where $\mu_w$ and $\sigma_w$ are the mean and standard deviation of $f_w$ respectively.

### 3.7 Multi-term indices

Changes in the relative importance of a single n-gram may be due to various reasons, as words may have different meanings in different contexts, but if we carefully select a set of words that are all related to the same concept, we may be interested in coordinated changes of all these words. This can range from the case when there are two alternative spellings for a concept, to its extreme where entire lexica of words are used for sentiment analysis (Pennebaker, et al., 2001). Making use of multi-term indices allows one to additionally test hypotheses based on pre-compiled lists of words that have been generated independently of the data, thus helping to avoid the risks that can be associated with multiple testing, and finding spurious signals in large datasets generally (Lansdall-Welfare, et al., 2017).

The multi-term index $g_W$ for a set of n-grams $W = \{w_1, w_2, \cdots, w_q\}$ is computed as

$$g_W(t) = \frac{\sum_{i=0}^{q} z_i(t)}{q}$$

where $q$ is the number of n-grams in the multi-term index and $z_i(t)$ is the value of the standardized time series of the $i^{\text{th}}$ element of $W$ at time $t$.

When using the multi-term index option, it is important to remember that as discussed in Section 3.6 on standardization, the magnitude of the underlying n-gram time series will be lost, as the average of the standardized time series is used to create the index. Additionally, consideration for the size of the confidence interval for each n-gram in the index should be taken into account. For example, if an n-gram is very infrequent, meaning we have a poor estimate of its relative frequency, *i.e.* the confidence interval is larger than the changes in the signal, then using it in a multi-term index can amplify the noise associated with this n-gram, affecting the overall trajectory of the set.

*3.8 Group change-points*

When analysing many different n-gram time series, it can be useful to not only look at their overall trend and the comparison between them, but also to investigate if there are times when they all experience a change together. These so-called group change-points represent points when a group of time series all change simultaneously, rather than each time series experiencing a change-point individually. Within the tool, these can be displayed with the group change-points option.

Group change-points are computed using the fast group LARS algorithm (Bleakley & Vert, 2011), which computes a piecewise constant approximation of each time series with common connected regions. Specifically, the time series are reconstructed under the constraint that sparse successive differences are cancelled group-wise in time (Bleakley & Vert, 2011) (Tibshirani, et al., 2005). Formally, for $F$ a set of $m$ n-gram time series of length $l$, the group change-points can be expressed as the minimization of $A$ in the convex optimization problem

$$\min_{A \in \mathbb{R}^{l \times m}} \frac{1}{2} \|F - A\|_2^2 + \lambda \sum_{i=1}^{l-1} \delta_i \|A_{i+1,\blacksquare} - A_{i,\blacksquare}\|_2$$

where $A_{i,\blacksquare}$ denotes the $i^{\text{th}}$ row of $A$, $\lambda$ penalises the group total variation and $\delta_i > 0$ is a position dependent correction used to alleviate some boundary effects. This is then further fine-tuned using dynamic programming as described further in (Bleakley & Vert, 2011).

## 4 Case Study on American Historical News

As an introduction to how one can start to use the tool to study the past, and to demonstrate the different options in the tool, we present a case study on the publicly available historical American news corpus from the Chronicling America project. This corpus consists of 8.9 million pages of news articles from over 1600 newspapers published between 1836 and 1922, selected by the project to best represent the different states' regional history, geographical coverage and the events that happened during the 87-year period of interest (Library of Congress and National Endowment for the Humanities, 2017). This corpus can be queried within the History Playground by selecting the "Chronicling America Archive (CAA)" option in the archive selection.

We decided to focus on the topic of sports within the corpus, using the documented topics reported by the Chronicling America project as a starting point[3]. For example, one of the sports-related topics is about the rise in popularity of basketball, where it states about the early decades of basketball in the 19th century that

*"Despite its humble beginnings in a YMCA gymnasium, Basket Ball expands more rapidly than any American sport in history. Over the span of twenty years, peach baskets are swapped with nets and backboards, the art of the dribble is near-perfected, and Dr. James Naismith becomes a household name."* (Library of Congress, 2014)

Among the other information presented is a list of important dates including the first official basketball contest in 1893, its rising popularity with women and co-ed organisations in 1894 due to its non-contact nature, and a shift in 1910 from basketball being a women's game to a serious intercollegiate men's contest that threatens football and baseball's position as the undisputedly most-beloved sport in America (Library of Congress, 2014). Using the Playground, we can begin to unpack each of these statements and gain an understanding of the data in the corpus supporting them.

*4.1 When did basketball begin coverage in the news?*

---

[3] Topics in Chronicling America available at https://www.loc.gov/rr/news/topics/topicsAlpha.html

Querying within the Playground for both "basketball" (which encompasses the alternative spelling of "basket-ball") and "basket ball", as shown in Fig. 3, we can immediately see that mentions of basketball in the corpus do indeed begin around 1893, with a minor preference for the 2-gram version of the spelling ("basket ball") over "basketball" or "basket-ball" for the first 12 years before eventually settling on the modern spelling as a single word.

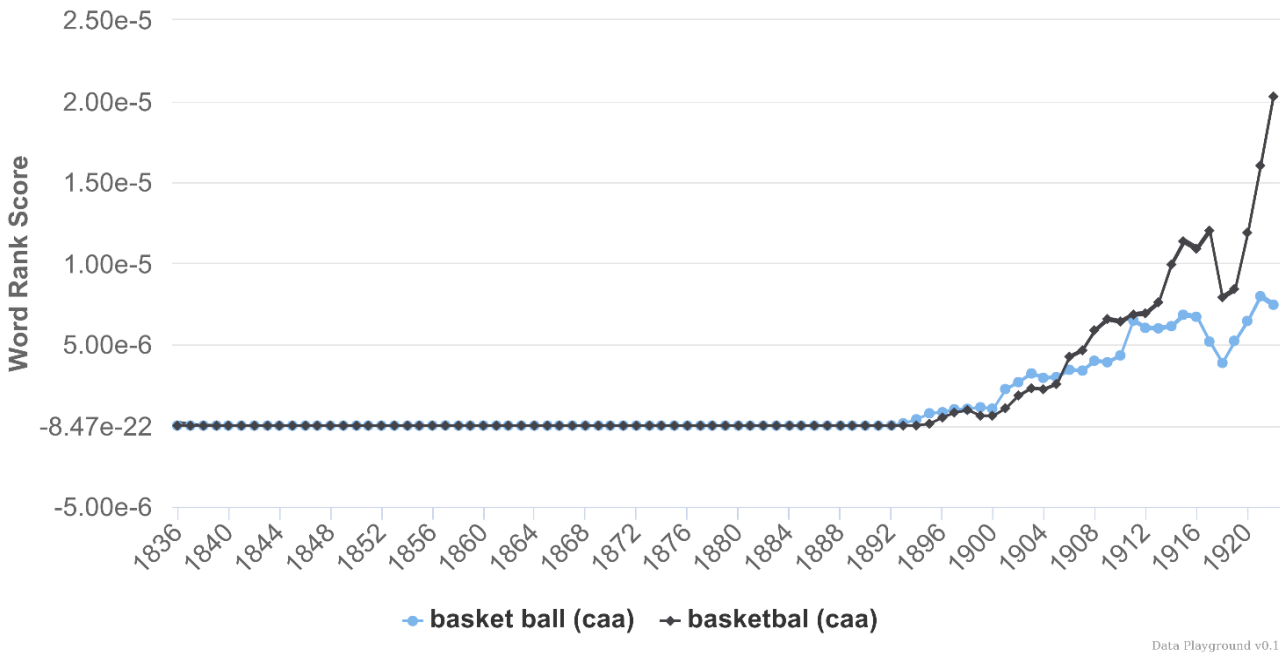

*Fig. 3 - Word rank score for the occurrences of "basket ball" (blue) and "basketball" (black) in the Chronicling America Archive (CAA).*

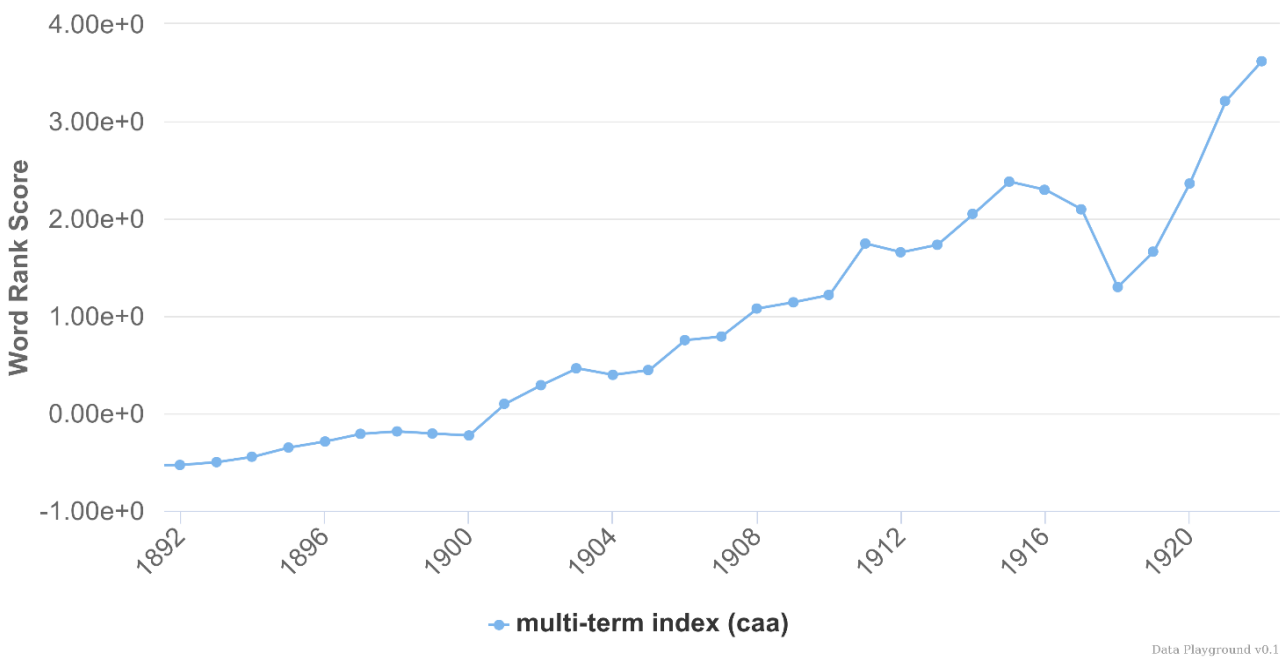

*Fig. 4 - Word rank score for the multi-term index composed of "basket ball" and "basketball" zoomed in to the 30-year period between 1892 and 1922 in the Chronicling America Archive (CAA).*

Indeed, using the multi-term index to combine both terms into one (Fig. 4), we can see that basketball continues to grow in prominence in the newspapers right up to the end of the period covered in the corpus in 1922, with a dip in 1918 during the first world war.

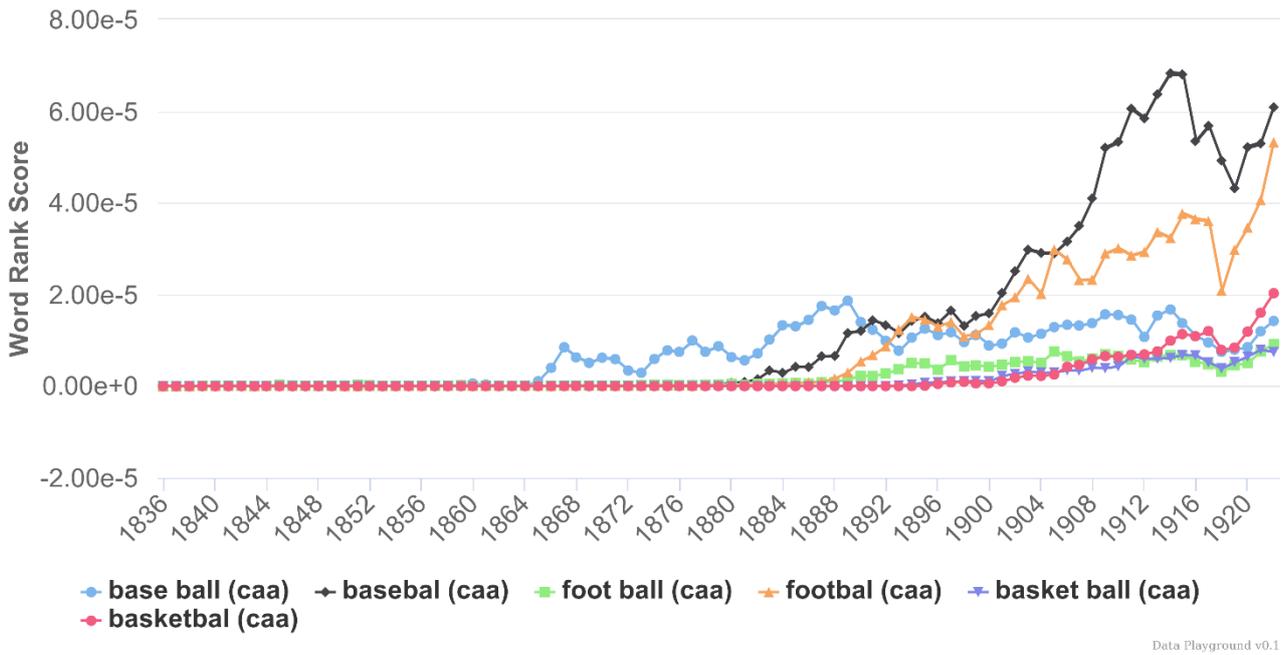

*Fig. 5 - Word rank score for the n-grams "base ball" (blue), "baseball" (black), "foot ball" (green), "football" (orange), "basket ball" (purple) and "basketball" (pink) in the Chronicling America Archive (CAA).*

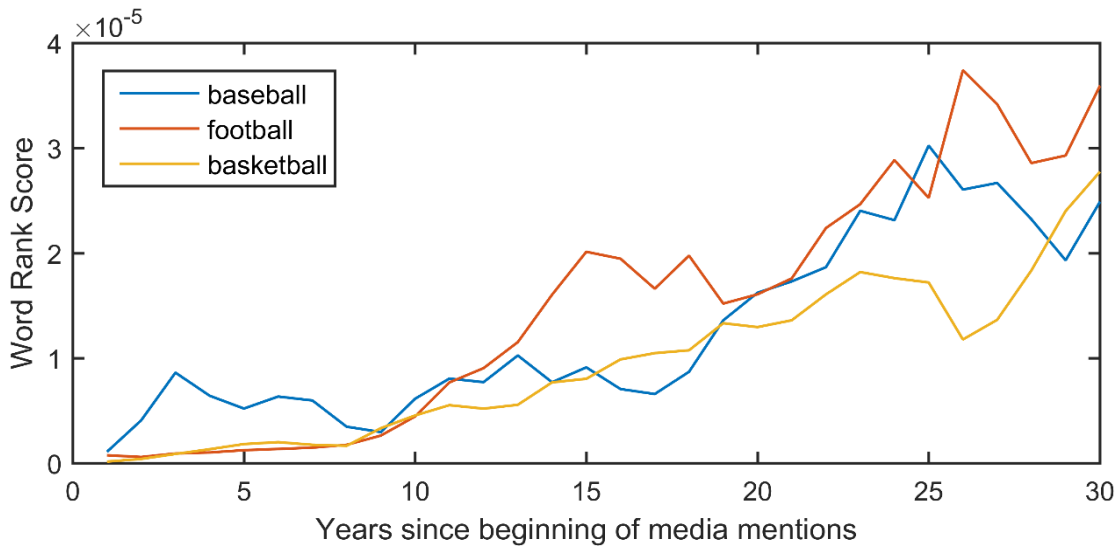

*Fig. 6 - Word rank score for baseball, football and basketball aligned to their first year of media mentions in the Chronicling America Archive (CAA).*

### 4.2 Does a shift occur around 1910?

While there is an increase in mentions in 1911, at this point it is still difficult to conclude whether basketball ever rises to the level of prominence experienced by baseball or football, or whether a shift occurred around this time.

By using the group-change points option of the Playground, we can find if, and when, any change points occurred within the time series. Applying this to the multi-term index time series shown in Fig. 4, we find that there is a change point in 1902, and again in 1911, lending support to the idea that a shift did occur around 1910.

*4.3 Does basketball rival baseball and football in prominence? Is it the fastest expanding American sport in history?*

Plotting the additional two n-grams of "baseball" and "football" (along with their alternative 2-gram spellings) (Fig. 5), we can see that baseball was originally picked up by the media in 1865, with the alternative one word spelling starting in 1880, followed by football beginning to be mentioned around 1883. It is interesting to see that, for all three sports, they originally began as two words, but later were blended into their modern-day one word versions.

Using the download CSV (comma-separated variable) option of the menu in the top right-hand corner of the chart, we can import the data into the statistical software package of choice to more closely examine the statements that *"Basket Ball expands more rapidly than any American sport in history"* and that *"[Basketball] threatens football and baseball's position as the undisputedly most-beloved sport in America"* (Library of Congress, 2014). This is performed by aligning the dates from which each sport is first mentioned in the media and compare their rates of growth.

By combining the time series of the different spellings for each sport and aligning them to the first year in which they begin to be mentioned, we can see in Fig. 6 that both football and basketball started with very similar beginnings in the media for the first 10 years. This differs to baseball, which enjoyed a large coverage in the first three years, which was then not matched again for a further 10 years. Overall, we find that basketball does appear to rival football and baseball in prominence within the media, at least in the first 30 years available here, but we do not find evidence that it is the fastest expanding American sport in history, challenging the view that *"the game's instant popularity has been unmatched in the history of sport"* (Spark, 2016).

*4.4 Additional notes*

We feel that while the case study provides some good examples and guidance on how the Playground can be used, and the types of conclusions that one can quickly draw from the data, there are various key design choices and their associated risks to consider when choosing n-grams to investigate. These include the ambiguity of an n-gram, how well it represents the concept under investigation, whether the meaning of the n-gram stays constant over time and many other factors, as further detailed in (Lansdall-Welfare, et al., 2017) (Nicholson, 2012).

Besides these choices, for the sake of demonstration, we have assumed in this case study that the relative frequency in the corpus can be used as a proxy for the popularity of a sport at a given time. While we believe that this is a fairly reasonable assumption in this case given the large number of newspaper outlets throughout the United States which are included and the public nature of sports and their reporting in general. However, it should be kept in mind that this, and other similar studies, are reflections of the underlying data. A more nuanced study may need to account for the readership and demographics of each newspaper, among several other factors depending on the corpus or corpora being analysed. A more in-depth discussion on this for the Google books corpus can be found in (Pechenick, et al., 2015).

## 5 Consideration for additional corpora

The History Playground has been designed to be able to incorporate many different corpora, all of which can be configured depending on the individual requirements.

While the examples shown here have been demonstrated on a corpus of English language text, the Playground can easily include corpora in other languages. For full support, languages also need to have available stemming algorithms, either in the Snowball package (Porter, 2001) or otherwise, however those not included can be supported without stemming.

Setting the resolution of a corpus is also an adjustable configuration for a corpus, allowing the tool to display and work with data that is not only over long time spans such as yearly data as in the case study, but also at hourly, daily, monthly or quarterly intervals, giving the option to study contemporary data such as social media or online news using the Playground, as demonstrated in an earlier prototype shown in (Lansdall-Welfare, 2015, pp. 139-150).

## 6 Discussion

As studied in (O'Sullivan, et al., 2015), collaboration between traditional humanities scholars and computer scientists is central to the field of digital humanities, with "clear acknowledgement that collaborators from the technical disciplines need to be treated more like stakeholders than enablers".

The benefits of analysing massive collections of text documents as a whole are becoming more apparent, with high profile studies (Leetaru & Schrodt, 2013) (Michel, et al., 2011) (Lansdall-Welfare, et al., 2017) (Dexter, et al., 2017) and media coverage bringing interest from a range of scholarly domains that work with text. And so, we have attempted to address the question of what tools exist and who will create new ones for researchers whom increasingly need tools that allow them access to large scale corpora and the ability to visualise them (Cheney, 2013), while addressing the concerns raised in (Nicholson, 2012) about the restrictive interfaces of online archives that currently make large-scale computational studies difficult.

Of course, one should be careful not to present such approaches and tools as simply visualisations of a corpus, as this misrepresents the amount of data processing, annotation, analysis and statistical testing that are involved in creating such interactive interfaces or conducting studies in this area. Tools like the History Playground for exploring massive textual corpora solve the myriad challenges one would otherwise have to tackle in order to perform these studies, including but not limited to the scalable collection, storage and annotation of vast amounts of text data, the extraction of meaningful information from the data, and visualising the data so that investigation into macroscopic patterns and trends within the data can be used to form and test hypotheses about the real world.

In general, these computational approaches work in complement with traditional close reading methods and require an interdisciplinary methodology that will continue to be necessary for the time being. This sentiment is mirrored in the work of other scholars who are actively developing tools (Klein, et al., 2015), where they state, as is also true in our case, that *"[Their work] is premised on the assumption that what is needed to advance scholarship in the digital humanities is not merely a new algorithm that confirms prior humanities research, nor a new humanistic domain for the application of existing algorithms. Rather, [they] seek to address the higher-level question of what can be gained from bringing together the humanities with computation through interdisciplinary collaboration."*

As these computational methods gain a critical mass, their reach will extend far into the digital humanities, political sciences and sociology, amongst others, shaping the types of studies that are possible, and to quote (Michel, et al., 2011) *"will furnish a great cache of bones from which to reconstruct the skeleton of a new science"*.

The hope is that in the long term, as more large textual datasets are released and additional feedback from the community helps to improve the Playground, we will be able to incorporate more varied and interesting corpora in the tool, along with continuing to develop analysis methods which can be applied to the data, and additional views and visualisations to help the user test their hypotheses and make sense of the data.

### Acknowledgments

Thomas Lansdall-Welfare and Nello Cristianini are supported by the ERC Advanced Grant "ThinkBig". The authors would like to thank Gaetano Dato for his helpful feedback and comments after using the tool, FindMyPast Newspapers Archive Ltd (www.britishnewspaperarchive.co.uk) for sharing the data with us and the Library of Congress for making their data available via an API.